\documentclass[conference]{ieeeconf}
\IEEEoverridecommandlockouts

\usepackage{amsmath,amssymb,amsfonts}
\usepackage{graphicx}
\usepackage{textcomp}
\usepackage{xcolor}
\usepackage{cite}
\usepackage{booktabs}
\usepackage{multirow}
\usepackage{array}
\usepackage[utf8]{inputenc}
\usepackage[T1]{fontenc}
\usepackage{url}
\usepackage{caption}
\graphicspath{{images/}}
\usepackage[
    colorlinks=true,
    linkcolor=blue,
    citecolor=red,
    urlcolor=blue,
    hypertexnames=false
]{hyperref}

\begin{document}

\title{Bio-Skin-Palm: An Open-Source Multimodal Tactile Sensor for the Palm of Dexterous Robotic Hands}

\author{Haoran Guo$^{1}$, Sicheng Xiong$^{1}$,
Ruilin Zhang$^{1}$, Haoyang Wang$^{2}$, and Chenxi Xiao$^{1*}$%
\thanks{$^{1}$H. Guo, S. Xiong, R. Zhang and C. Xiao are with ShanghaiTech Univeristy, Rimalb
        (email: guohr2026, xiaochx@shanghaitech.edu.cn).}%
\thanks{$^{2}$H. Wang is with the University of Alberta
        (e-mail: haoyan25@ualberta.ca).}%
\thanks{*Corresponding Author}%
}

\maketitle

\noindent\begin{abstract}
Tactile sensing is critical for dexterous manipulation, yet most robotic hands remain largely fingertip-centric, leaving the palm, the hand’s largest continuous interaction surface, unsensed. Our insight is that palm sensing should complement rather than replicate fingertip sensing by providing multimodal perception throughout the interaction lifecycle: pre-contact, contact, and post-contact. We present \emph{Bio-Skin-Palm}, an open-source, cost-effective multimodal palm sensor for dexterous hands. Its 5.1-mm-thick layered structure integrates four sensing modalities: fifteen single-axis Hall channels for distributed normal force, a triaxial Hall sensor for global shear force, two time-of-flight arrays for pre-contact proximity and coarse geometry, and a heater--thermistor pair for active thermal perception. To improve electromagnetic robustness, a permalloy layer passively homogenizes external magnetic fields, while an overlaid force-sensitive resistor layer enables active compensation of residual Hall-sensor bias. The resulting sensor provides 0--5~N measurement ranges for both normal and shear force. Passive shielding reduces static offset and dynamic RMS error up to 36.7\%, while active compensation reduces Hall-sensor RMSE by about 75\%. Integrated with a LEAP Hand, Bio-Skin-Palm enables RGB-free teleoperated retrieval from an opaque duffel bag and thermal material discrimination, both achieving over 88\% success. An autonomous closed-loop experiment further demonstrates its potential for long-horizon manipulation involving heterogeneous sensing modalities. With a fabrication cost below 30 \,USD, Bio-Skin-Palm provides an affordable and open-source platform for multimodal palm sensing in dexterous manipulation.
Supplementary Project Page:\newline
{\urlstyle{same}\nolinkurl{https://anonymoussubmit-6677.github.io/Bio-Skin-Palm/}}
\end{abstract}

\section{Introduction}

\noindent Tactile sensing is fundamental to physical interaction in both biological and robotic hands~\cite{li2026biomimetic}. Beyond detecting contact and characterizing force, tactile sensing can provide complementary physical information, including proximity, geometry, temperature, and surface properties, supporting grasp stabilization, slip detection, blind grasping, and manipulation under unreliable visual perception~\cite{zhou2026activepalm,james2018slip,lee2026blindgrasp,luo2026blindgrasp,mao2024housekeeping}. However, tactile sensing on robotic hands remains predominantly fingertip-centric~\cite{allegrov5,lambeta2020digit}. In contrast, the palm has received considerably less attention, despite being the hand's largest continuous interaction surface and playing a central role in object enclosure and grasp support~\cite{zhang2024tacpalm,su2024inflatable}.

\begin{figure}[!t]
\centering
\vspace{6pt}
\includegraphics[width=0.96\columnwidth]{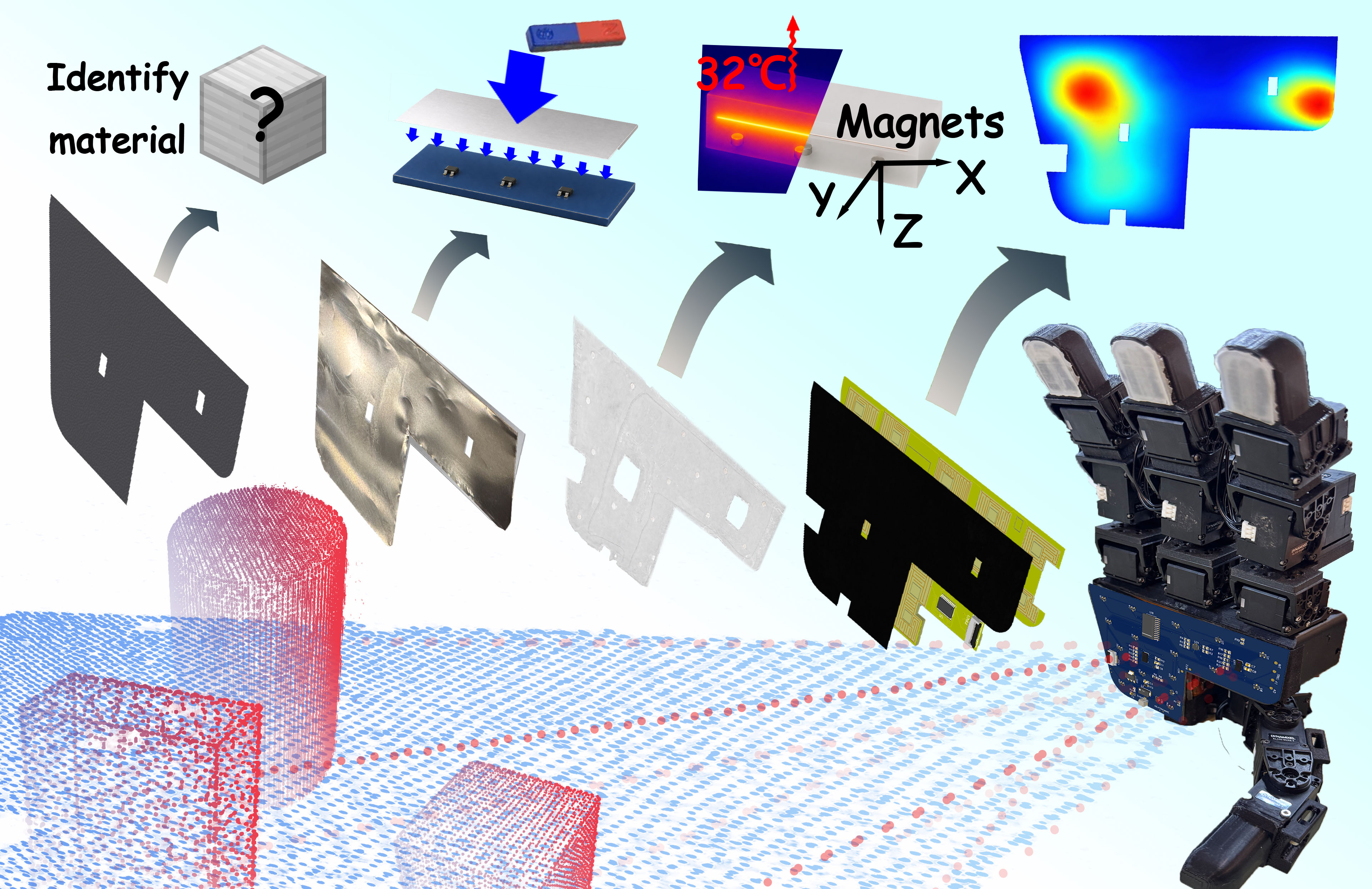}
\vspace{-4pt}
\caption{Overview of the \textbf{multimodal tactile palm} integrated into a LEAP Hand. The system combines material recognition, active interference compensation, normal force sensing, shear force sensing, and proximity perception of object position and coarse shape. Left to right: thermal sensing, magnetic-shielding, heater--magnet, FSR, and PCB base layers.}
\vspace{-18pt}

\label{fig}
\end{figure}

Existing tactile technologies do not yet provide a practical solution for multimodal palm sensing. Research prototypes have explored either optically rich sensing systems that are bulky, expensive, and difficult to integrate into the palm~\cite{yuan2017gelsight,zhang2024tacpalm,zhou2026activepalm} or dense force arrays that provide only a single sensing modality~\cite{li2024palmteng,su2024inflatable}. In commercial robotic hands, tactile channels are typically concentrated at the fingertips~\cite{allegrov5,sharpawave}, while palm sensing is relatively absent. Recent study begun to explore palm sensing for dexterous manipulation, but found that adding only force-based tactile information to the palm provides limited gains for learning-based manipulation models~\cite{guo2024sensorconfig}. This finding does not imply that palm sensing is unnecessary; rather, it raises a more fundamental design question: \emph{what information should the palm sense that fingertips cannot provide as effectively?}

We argue that the answer lies in exploiting the palm's unique spatial and temporal role during interaction. Unlike finger joints, whose contact angles vary during grasping, the palm can remain oriented toward the object throughout a manipulation episode. This makes it particularly suitable for \emph{staged multimodal perception}. \textbf{Before contact}, palm-scale proximity sensing can localize an approaching object and provide coarse geometric information~\cite{hsiao2009proximity,dufrene2026tof}. \textbf{Contacting}, distributed force sensing in the palm can detect contact with the target object instantly, prevent damage from excessive transient forces, and help determine the appropriate time to grasp. \textbf{After contact}, the palm can maintain a relatively large and mechanically stable contact area, enabling thermal measurements that require sustained contact and controlled heat transfer. Crucially, the sensing modalities are not restricted to specific stages. They remain available throughout, with some modalities serving supporting roles at each stage. Thus, the opportunity is not to extend fingertip sensing over a larger area, but to use the palm as a persistent sensing interface that provides different information at each stage.

Motivated by this perspective, we present \emph{Bio-Skin-Palm}, an open-source, cost-effective multimodal tactile palm designed in functional layer structure (Fig.~\ref{fig}). The 5.1mm thickness integrates fifteen single-axis Hall channels for distributed normal-force sensing, one triaxial Hall sensor for global shear-force estimation, twelve FSR zones for pressure sensing and cross-modal compensation, two $8\times8$ ToF sensors for pre-contact perception, and a heater--thermistor pair for active thermal sensing. To improve robustness against magnetic interference, a permalloy layer passively homogenizes external magnetic flux, while an FSR-gated compensator actively corrects residual Hall bias. On independent test data, the Hall and FSR arrays achieve normal-force MAEs of 0.246N over 0-5N and 0.799N over 0--10N, respectively. The shear estimator achieves a magnitude MAE of 0.259N with a mean directional error of $4.00^\circ$. Passive shielding reduces static offset and dynamic RMS variation by 36.7\% and 33.2\%, respectively, while active compensation reduces interference-induced RMSE by 74.99\%. Integrated with a LEAP Hand, the system supports RGB-free object identification and retrieval, thermal based material discrimination. Moreover, a closed-loop experiment shows that a universal VLM, running locally without tuning, can utilize different sensing modalities to complete a long-horizon hybrid task involving both geometric and material objectives, providing strong evidence that our multimodal palm can offer decision-critical information at each stage of manipulation.

The main contributions are as follows. Hardware, firmware, software, and a fabrication tutorial will be fully released as \textbf{Open Source} upon publication.

\begin{enumerate}
    \item \textbf{Multimodal tactile palm.} An open-source, 5.1mm-thick palm that integrates pre-contact proximity sensing, multidirectional contact force sensing, and post-contact thermal sensing in a layered architecture.
    
    \item \textbf{Hybrid magnetic-interference mitigation.} A combination of passive permalloy flux homogenization and active FSR-gated cross-modal compensation that improves the robustness of Hall-based force sensing.
    
    \item \textbf{System-level multimodal validation.} Comprehensive validation of both individual sensing modalities and their system-level integration, including RGB-free object identification and retrieval, thermal-based material discrimination, and a long-horizon closed-loop task in which a locally deployed universal VLM uses geometric and material cues without fine-tuning.
\end{enumerate}

\section{Related Work}

\subsection{Tactile Sensors for the Robotic Palm}

\noindent Palm sensing extends tactile perception beyond the fingertips to the palm contact region involved in object enclosure and load sharing. Its larger surface area also provides greater flexibility for integrating distributed and heterogeneous sensing modalities. Recent systems have explored inflatable optical palms for surface reconstruction~\cite{su2024inflatable}, visual--tactile and active palms for palm--finger coordination and contact-rich manipulation~\cite{zhang2025palmfinger,zhou2026activepalm,zhang2024tacpalm}, and triboelectric arrays for force sensing~\cite{li2024palmteng}. Commercial robotic hands likewise incorporate dense palm force arrays~\cite{allegrov5,sharpawave}. However, these systems typically either rely on costly designs, or adopt designs whose functionality is largely limited to conventional multipoint force sensing. Compact, low-cost, and open-source palm sensors that integrate complementary multimodal sensing remain underexplored.

\subsection{Multimodal Tactile Sensing for Robotic Hands}

\noindent Biological touch integrates mechanical and thermal cues to infer contact, temperature, and surface properties. Inspired by this capability, robotic tactile systems have combined force with temperature and
proximity~\cite{li2026biomimetic}, pressure with magnetic-field or
range sensing~\cite{song2025fabric,li2025proximitytactile}, and tactile sensing with vision~\cite{mao2024housekeeping}. The MOTIF Hand integrates tactile, depth, thermal, inertial, and visual sensing for temperature-aware grasping, However, these sensing modalities are distributed across the entire hand but not integrated into the palm~\cite{zhou2025motif}. These systems demonstrate the value of multimodal physical perception, but often offer limited integration, with different modalities distributed across separate components, and remain relatively high price. As summarized in Table~\ref{tab:comparison}, Bio-Skin-Palm addresses this gap with a compact, cost-effective, and open-source architecture that integrates proximity, force, and thermal sensing within a thin palm, enabling complementary physical perception before, during, and after contact.

\subsection{Manipulation under Limited Visibility}

\noindent Manipulation under darkness or occlusion requires embodied sensing to compensate for unreliable visual observations. Existing vision-free approaches primarily infer object states from fingertip tactile and proprioceptive feedback acquired after contact~\cite{lee2026blindgrasp,luo2026blindgrasp}. Although learned policies can act on these signals, their situational awareness remains constrained by the local and reactive nature of fingertip contact sensing. Bio-Skin-Palm addresses this limitation at the sensing-hardware level by combining palm-mounted ToF, force, and thermal sensing, providing complementary physical information before, during, and after contact and thereby extending tactile perception beyond local contact events.

\begin{table}[t]
\centering
\caption{Comparison with representative tactile designs. A checkmark indicates
a supported sensing modality. ``Site'' denotes the tactile-array location;
$t$ is the stack thickness; N/R means not reported.
$^\dagger$Whole-hand price; the remaining values are sensor costs. 
$^{\ddagger}$TacPalm infers grasp pose from contact; ours from pre-touch proximity.}
\vspace{-4pt}
\label{tab:comparison}
\scriptsize
\setlength{\tabcolsep}{2pt}
\renewcommand{\arraystretch}{1.1}

\resizebox{\columnwidth}{!}{%
\begin{tabular}{lcccccc}
\toprule
 & Allegro V5 & SharpaWave & TacPalm & Infl.\ Palm & Bio-Skin & \textbf{Ours} \\
 & \cite{allegrov5} & \cite{sharpawave} & \cite{zhang2024tacpalm}
 & \cite{su2024inflatable}
 & \cite{guo2025bioskin} & \\
\midrule
Normal force & $\checkmark$ & $\checkmark$ & $\checkmark$ & $\checkmark$
             & $\checkmark$ & \textbf{$\checkmark$} \\
Shear force  & -- & $\checkmark$ & -- & --
             & $\checkmark$ & \textbf{$\checkmark$} \\
Thermal      & -- & -- & -- & --
             & $\checkmark$ & \textbf{$\checkmark$} \\
Proximity    & -- & -- & -- & --
             & -- & \textbf{$\checkmark$} \\
Grasp Pose         & -- & -- & $\checkmark$ & --
             & -- & $\checkmark^{\ddagger}$ \\
\midrule
Site         & hand & tip & palm & palm & tip & \textbf{palm} \\
$t$ (mm)     & N/R & N/R & 5.0 & N/R & 15.0 & \textbf{5.1} \\
Cost (\$)    & $>$10,000$^\dagger$ & $\sim$50,000$^\dagger$ & N/R & N/R
             & 10 & \textbf{30} \\
Rate (Hz)    & N/R & 180 & 30 & N/R & 200 & $\leq $\textbf{500}
 \\
Open source  & No & No & Yes & N/R & Yes & \textbf{Yes} \\
\bottomrule
\end{tabular}%
}
\vspace{-12pt}
\end{table}

\section{Sensor Design}

\noindent
Bio-Skin-Palm follows the design philosophy of Bio-Skin~\cite{guo2025bioskin}, while introducing key advances in three aspects (Table~\ref{tab:comparison}). First, we largely reduce thickness and enable palm integration by replacing Bio-Skin's side-wall FSR shear force sensors with a single triaxial Hall sensor, significantly reducing the sensor stack thickness from 15.0\,mm to 5.1\,mm. Second, we improve robustness to external magnetic disturbances through passive shielding and active compensation (Sec.~\ref{sec:emi}). Third, we incorporate proximity sensing to enable pre-contact perception. An overview of the proposed system is shown in Fig.~\ref{fig:architecture}. The BOM is available on the project page.

\subsection{Sensor Modality and Architecture}

\begin{figure}[t]
    \centering
    \includegraphics[width=0.99\linewidth]{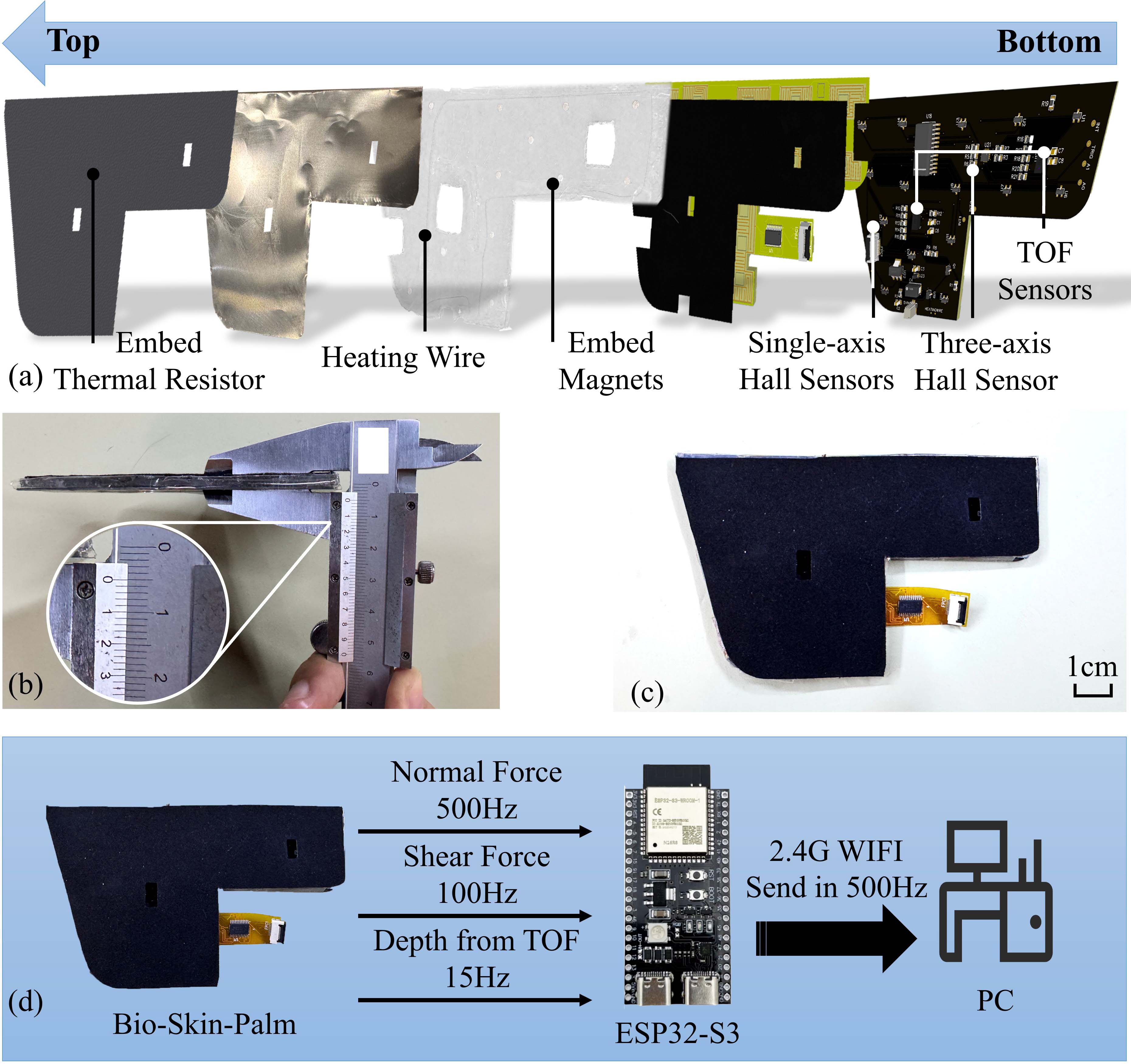}
\vspace{-16pt}
\caption{Multimodal tactile-palm hardware and acquisition system. 
(a)~Five-layer structure integrating thermal sensing and heating, 
normal/shear-force sensing, and 
ToF depth perception. 
(b)~The assembled stack is only 5.1\,mm thick. 
(c)~Palm-shaped sensor integrated with the LEAP Hand. 
(d)~Wireless acquisition of normal force (500\,Hz), shear force (100\,Hz), and depth (15\,Hz), unified into a 500\,Hz multimodal stream over 2.4\,GHz Wi-Fi.}
\vspace{-4pt}
\label{fig:architecture}
\vspace{-12pt}
\end{figure}

\noindent Overall, the sensor comprises five functional layers (Fig.~\ref{fig:architecture}a): the PCB substrate, FSR layer, heating-and-magnet layer, permalloy shielding layer, and temperature-sensing layer.

\noindent \textbf{(1) PCB substrate.}
The PCB integrates multiple sensing units. First, two $8\times8$ ToF modules (VL53L5CXV0GC) perceives object geometry (Sec.~\ref{sec:teleop}). Second, fifteen distributed DRV5056A3QDBZT single-axis Hall sensors measure normal-direction magnetic-field changes to infer the normal-force distribution. Third, one MLX90393 triaxial Hall sensor measures tangential magnetic-field changes to estimate a global shear-force vector.

\noindent \textbf{(2) Force-sensitive resistor (FSR) layer.}
The FSR layer is a custom flexible printed circuit coated with a flexible resistive film, which covers the palm in twelve sensing zones. Rather than replacing the Hall array, the FSR layer serves two complementary roles. First, it provides an independent reference for detecting and compensating magnetic disturbances: external magnetic fields can bias Hall measurements even under no load, whereas the piezoresistive response is insensitive to such disturbances. Second, it extends the measurable force range beyond that of the Hall array. However, forces below $0.5\,\mathrm{N}$ may fall below the detection threshold of the FSR layer and therefore still rely on the Hall sensors. Hence, the Hall array remains the primary sensor for normal-force measurement, with the FSR layer providing disturbance compensation and force-range extension.

\noindent \textbf{(3) Magnet and heating layer.}
This layer embeds sixteen small cylindrical magnets within a compliant silicone elastomer, aligned with the sixteen Hall sensing locations. External loads displace each magnet relative to its corresponding Hall sensor, producing magnetic-field variations that are converted into normal-force measurements~\cite{yan2021magskin,bhirangi2021reskin}. A heating wire is also embedded in the silicone elastomer to regulate the sensor temperature. With closed-loop temperature control, the heating element maintains the sensor surface near $31^{\circ}\mathrm{C}$, approximately matching human skin temperature.

\noindent \textbf{(4) Permalloy shielding layer.}
The permalloy layer provides magnetic shielding through its high magnetic permeability (at least $280\,\mathrm{mH/m}$). When exposed to a strong, localized external magnetic field, we observed that the layer redistributes the magnetic flux across neighboring channels, converting a large local error into a smaller and more spatially uniform perturbation. This reduces the peak disturbance in individual Hall channels and improves the robustness of normal-force estimation against external magnetic interference.

\noindent \textbf{(5) Temperature-sensing layer.}
The temperature-sensing layer integrates negative-temperature-coefficient (NTC) thermistors for fast surface-temperature measurement, providing feedback for closed-loop thermal regulation.

\subsection{Embedded System and I/O Design}

\noindent An ESP32-S3 serves as the central controller for sensor coordination and data acquisition. To minimize external cabling that could interfere with hand motion, the system communicates wirelessly over 2.4\,GHz Wi-Fi. On the sensing side, the normal-force and temperature channels are sampled at 500\,Hz. In parallel, the shear-force sensor and the two ToF modules operate on independent I\textsuperscript{2}C buses at 100\,Hz and 15\,Hz, respectively. All sensor data are then broadcast as UDP frames at a fixed rate of 500\,Hz. For channels with lower sampling rates, the most recent measurement is retained using a zero-order hold.

\section{Multimodal Signal Processing}
\label{sec:basic}

\noindent The sensor outputs are processed into four physical representations: distributed normal force, global shear force, surface temperature, and two ToF depth images. The processing pipeline first detects and mitigates magnetic interference in the Hall measurements, and then converts the corrected signals into physical measurements. In parallel, the ToF measurements are independently reconstructed into 3D surface point clouds.

\subsection{FSR-gated magnetic-bias compensation (FG-MBC)}
\label{sec:fgmbc}

\begin{figure}[t]
    \centering
    \includegraphics[width=\linewidth]{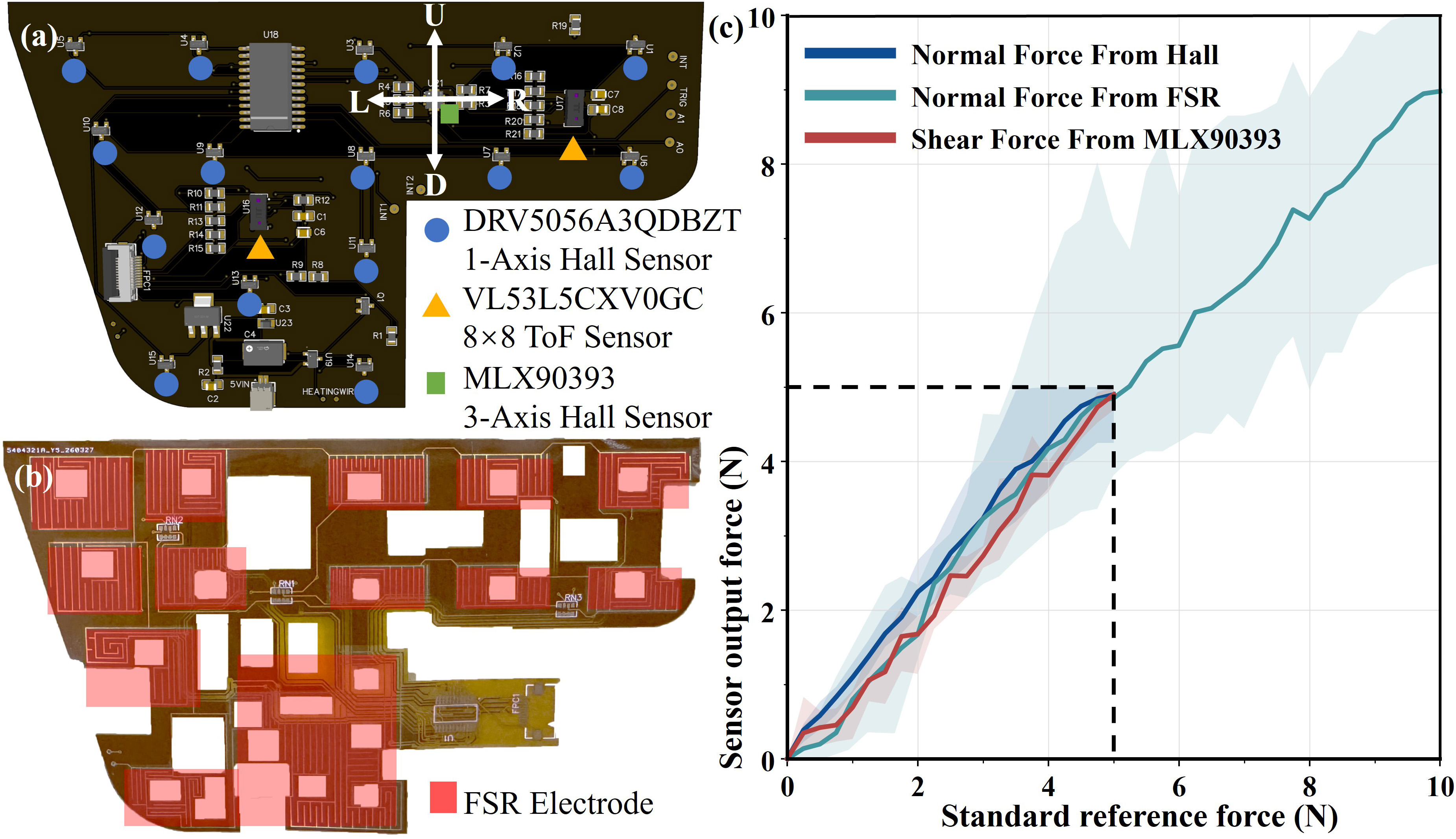}
    \vspace{-6pt}
    \caption{Force-channel layout and calibrated response.
    (a)~Fifteen single-axis Hall sensors (blue), the triaxial
    sensor (green) and calibration directions (white),
    two ToF modules (yellow).
    (b)~Twelve FSR zones.
    (c)~Calibrated force responses: channel means
    (solid curves) and full spread (shading).}
    \vspace{-15pt}
    \label{fig:characterization}
\end{figure}

\noindent External magnetic fields can produce false Hall-sensor responses without physical contact. To detect and compensate for such interference, the FSR over each Hall channel provides an independent contact reference, as shown in Fig.~\ref{fig:characterization}(a)(b). At step $k$, let $C_i=1$ denote FSR-detected contact and $I_i=1$ denote a Hall-force estimate above the worst-case FSR dead band of $0.5\,\mathrm{N}$. Hence, $C_i=0$ and $I_i=1$ indicate magnetic interference. Smaller responses are ignored because they lie within the FSR dead band and are not safety-critical.

For Hall channel $i$, let $A_i[k]$ be the raw ADC reading and $b_i$ its unloaded baseline. We define the positive deviation as
$e_i[k]=\max\{A_i[k]-b_i,0\}$ and estimate the magnetic bias $d_i[k]$ as
\begin{equation}
\small
d_i[k]=
\begin{cases}
d_i[k-1],
& C_i=1, \\[3pt]
(1-\alpha_a)d_i[k-1]+\alpha_a e_i[k],
& C_i=0,\ I_i=1, \\[3pt]
(1-\alpha_r)d_i[k-1],
& C_i=0,\ I_i=0.
\end{cases}
\label{eq:fgmbc}
\end{equation}

where $\alpha_a$ and $\alpha_r$ are the adaptation and decay rates. Thus, the bias is held during contact, updated under detected interference, and otherwise gradually decayed.

When FSR contact is detected, we keep the bias unchanged so that
force-induced changes in the Hall signal are preserved. When the Hall
signal is strong but the FSR detects no contact (suggesting external
magnetic interference), we update the bias toward the observed Hall
deviation to track the interference. In all other cases, the estimated
bias gradually decays toward zero. The corrected Hall reading is then
computed as $A_i^c[k]=A_i[k]-d_i[k]$ and converted to normal force. In
this way, the FSR provides a contact gate for adaptive bias estimation,
while the Hall array remains the primary sensor for sensitive
normal-force measurement.

\subsection{Force and Thermal Sensor Readout and Calibration}
\label{sec:calibration}

\noindent
After drift compensation, the calibrated force readings are obtained from the Hall and FSR sensors. Specifically, the fifteen Hall channels and twelve FSR zones cover force ranges of $0$--$5\,\mathrm{N}$ and $0$--$10\,\mathrm{N}$, respectively, while the multiaxis MLX90393 measures shear forces of up to $5\,\mathrm{N}$ along each of the four Left/Right/Up/Down directions. In addition, NTC readings are acquired for temperature measurement. All raw sensor readings are converted into physical quantities through the calibration procedures detailed in Appendix~\ref{app:forcecal} and ~\ref{app:thermalcal}.

\subsection{ToF Surface Reconstruction}
\label{sec:pointcloud}

\noindent
In addition to force and temperature, distance measurements from the ToF sensors are reconstructed into a voxelized surface representation. Specifically, the depth measurements are back-projected into 3D and transformed into the world frame. Measurements from the two ToF modules are then filtered and fused to remove inconsistent points. The resulting 3D points are accumulated into a voxelized surface cloud using $10\,\mathrm{mm}$ cubic voxels while scanning a static scene~\cite{hornung2013octomap}.

\section{Experiment}
\label{sec:experiment}

\noindent
We evaluate the proposed palm sensor across three stages of manipulation. (1) \textbf{Pre-contact}: ToF sensing for object localization (Sec.~\ref{sec:teleop}); (2) \textbf{Contact}: force sensing with magnetic shielding and interference compensation (Sec.~\ref{sec:emi}); and (3) \textbf{Post-contact}: thermal sensing for material recognition (Sec.~\ref{sec:material}). Finally, Sec.~\ref{sec:closedloop} evaluates the integration of these modalities in a single autonomous, long-horizon manipulation task involving three stages.

\subsection{Pre-Contact Stage: Vision-Degraded Teleoperation}
\label{sec:teleop}

\begin{figure}[t]
\centering
\includegraphics[width=0.95\columnwidth]{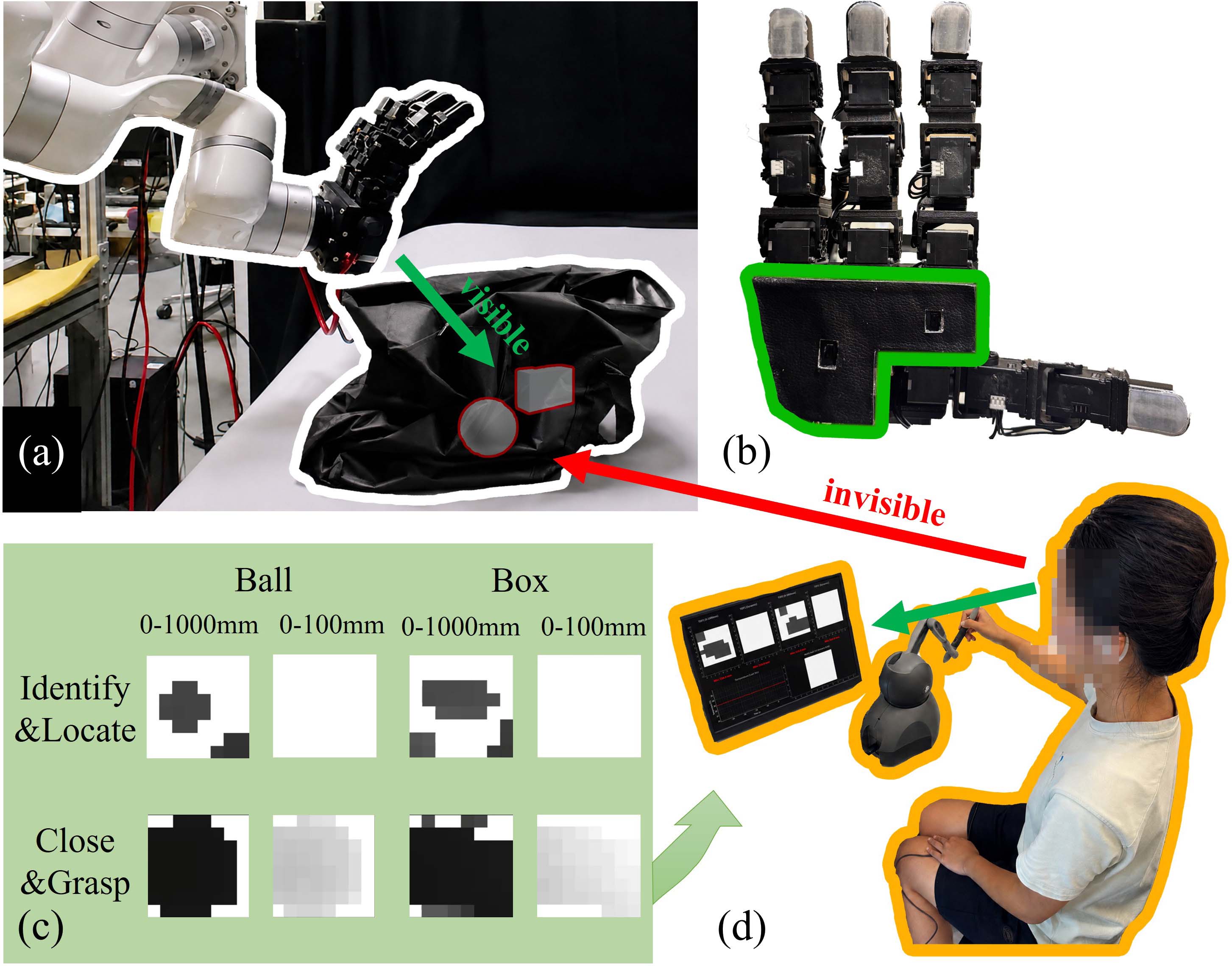}
\caption{Vision-degraded teleoperation.
\textbf{(a)} The LEAP Hand retrieves a sphere or box from an opaque bag without direct or external RGB-camera visibility.
\textbf{(b)} Bio-Skin-Palm mounted on the palm, showing its two ToF apertures.
\textbf{(c)} ToF depth maps used for object identification and grasp closure; darker cells indicate closer surfaces.
\textbf{(d)} Teleoperation using a Geomagic Touch haptic device, with the palm ToF display as the only visual feedback.}
\vspace{-15pt}
\label{fig:teleop}
\end{figure}

\noindent
Approaching an object is a prerequisite for manipulation, yet conventional RGB-based sensing can become unreliable under darkness and occlusion, such as inside a constrained bag. We therefore conducted a teleoperation human study to evaluate whether palm-mounted ToF sensing can support object search and approach when RGB vision is unavailable.

\textbf{Setup and protocol.}
As shown in Fig.~\ref{fig:teleop}, five consenting participants (aged 21-27, all males, all self-reported healthy without disability) used a Geomagic Touch haptic device in front them to control the Cartesian pose of an xArm robot. The arm carried a LEAP Hand~\cite{shaw2023leap} equipped with Bio-Skin-Palm, while buttons on the haptic device triggered preprogrammed grasp actions. The system was not equipped with an RGB camera; the only visual feedback available to participants was the $8\times8$ ToF data, displayed as two grayscale maps. The depth range was set to $0$--$1000\,\mathrm{mm}$ for object search and approach and $0$--$100\,\mathrm{mm}$ for determining grasp closure timing. 

A $5\times15\times15\,\mathrm{cm}$ box and a sphere with a diameter of $10\,\mathrm{cm}$ were randomly placed at non-adjacent positions inside an opaque black woven bag. After one minute of familiarization with the haptic device, each participant declared the target object before each attempt and attempted to grasp it until completing three successful retrievals per object. An attempt was considered successful if the participant correctly grasped the declared object and lifted it by $0.3\,\mathrm{m}$. Object positions were randomized after every attempt.

\textbf{Results.}
As shown in Table~\ref{tab:teleop}, participants correctly located and identified the declared target in all 34 attempts, achieving a 100\% target localization rate. They successfully lifted the target in 30 attempts, resulting in an overall lift success rate of 88.2\%, with individual success rates ranging from 66.7\% to 100\%. All four failed lifts were caused by grasp timing or approach orientation rather than target search or identification. These results demonstrate that the palm-mounted ToF sensing provided sufficient distance and coarse shape cues for participants to locate and approach the target inside the dark bag. However, participants reported difficulty distinguishing details between similar objects due to the limited spatial resolution. This limitation highlights the need for complementary sensing modalities. Accordingly, Sec.~\ref{sec:material} evaluates the thermal sensing in material recognition.

\subsection{Contact Stage: Force Accuracy and Magnetic Robustness}
\label{sec:emi}

\begin{figure}
    \centering
    \includegraphics[width=0.95\linewidth]{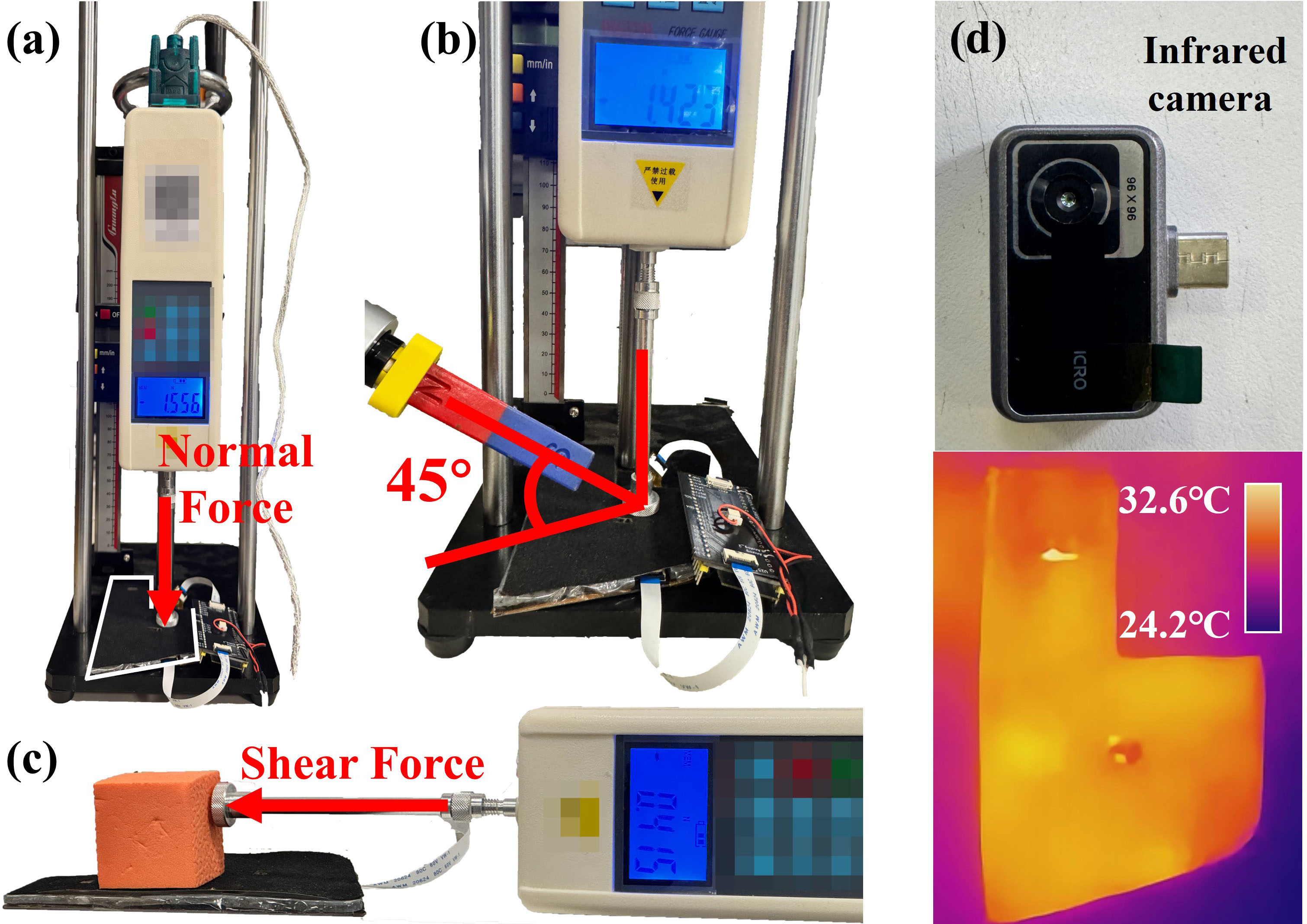}
    \caption{Force and temperature calibration process. (a) Normal force calibration. (b) FG-MBC test setup. (c) Shear force calibration. (d) Temperature calibration with infrared camera.}
    \vspace{-18pt}
    \label{fig:forceTempCali}
\end{figure}

\noindent This section characterizes the force-sensing performance of Bio-Skin-Palm. While Hall sensors enable sensitive multidirectional force measurements, they are susceptible to external magnetic fields. We therefore evaluate two aspects: (i) force accuracy under interference-free conditions, and (ii) robustness to magnetic interference. For accuracy characterization, we first validate the calibration models. We then evaluate passive permalloy shielding and active FG-MBC for mitigating magnetic interference. The devices used for calibration and FG-MBC evaluation are shown in Fig.~\ref{fig:forceTempCali}.

\begin{table}[t]
    \centering
    \caption{Vision-degraded teleoperation performance. Each participant completed three successful lifts per object. Lift success rate is the ratio of successful lifts to total grasp attempts.}
    \vspace{-4pt}
    \label{tab:teleop}
    \small
    \setlength{\tabcolsep}{4pt}
    \begin{tabular}{lcccc}
        \toprule
        & \multicolumn{2}{c}{Grasp attempts} & \multicolumn{2}{c}{Success rate (\%)} \\
        \cmidrule(lr){2-3}\cmidrule(lr){4-5}
        Participant & Box & Sphere & Target Localization & Lift \\
        \midrule
        P1 & 4 & 3 & 100 & 85.7 \\
        P2 & 5 & 4 & 100 & 66.7 \\
        P3 & 3 & 3 & 100 & 100  \\
        P4 & 3 & 3 & 100 & 100  \\
        P5 & 3 & 3 & 100 & 100 \\
        \midrule
        Overall & 18 & 16 & 100 & 88.2 \\
        \bottomrule
    \end{tabular}
    \vspace{-6pt}
\end{table}

\begin{table}[t]
    \caption{Calibrated force accuracy on held-out data. MAE and RMSE are the $L_1$
    and $L_2$ norms of $\hat F-F$; NMAE is normalized by each row's own scale and
    is \textbf{not comparable across rows}. Metrics are channel-averaged.}
    \vspace{-4pt}
    \label{tab:basic}
    \centering
    \scriptsize
    \setlength{\tabcolsep}{2.5pt}
    \renewcommand{\arraystretch}{1.1}
    \begin{tabular}{lcccccc}
        \toprule
        Modality & Range & MAE & RMSE & NMAE & $R^2$ & Hyst. \\
                 & (N)   & (N) & (N)  & (\% FS) &     & (\% FS) \\
        \midrule
        Hall array, $F_z$ (15 ch.) & 0--5  & 0.246 & 0.306 & 4.92 & 0.940 & 4.34 \\
        FSR array, $F_z$ (12 zones)& 0--10 & 0.799 & 0.993 & 7.99 & 0.853 & 4.75 \\
        Triaxial, $\lVert F_{xy}\rVert$ & 0--5 & 0.259 & 0.319 & 5.17 & 0.949 & 11.85 \\
        \bottomrule
    \end{tabular}
\vspace{-15pt}
\end{table}

\textbf{(1) Normal force.} To characterize the response from normal indentation, all channels were calibrated against a force gauge with the sensor temperature maintained at steady $31^{\circ}\mathrm{C}$. Table~\ref{tab:basic} summarizes the results, and Fig.~\ref{fig:characterization}(c) shows the corresponding calibration curves. The Hall array achieved an MAE of $0.246\,\mathrm{N}$ over $0$--$5\,\mathrm{N}$ (4.92\% FS, $R^{2}=0.940$), providing the primary normal-force measurement. The FSR array achieved an MAE of $0.799\,\mathrm{N}$ over $0$--$10\,\mathrm{N}$ (7.99\% FS, $R^{2}=0.853$), offering a wider measurement range and a magnetically insensitive reference for FG-MBC.

Because the magnets embedded in the gel may affect neighboring channels, we characterized the spatial response of normal indentation by loading a single cell. The maximum response among neighboring channels averaged 4.85\% FS for the Hall array and 2.78\% FS for the FSR array. Additional characterization sheets are provided on the project page.

\textbf{(2) Shear force.} The triaxial sensor achieved a shear-force MAE of $0.259\,\mathrm{N}$ over the $0$--$5\,\mathrm{N}$ range ($R^{2}=0.949$), with a mean force-direction error of $4.00^{\circ}$. The shear channels showed a loading--unloading hysteresis of $0.59\,\mathrm{N}$ ($11.85\%$ FS), which was larger than that of the normal-force channel.

\textbf{(3) Passive permalloy layer.} We evaluated the magnetic-shielding effect of the passive permalloy layer without FG-MBC compensation. An external magnetic interferer was placed $2\,\mathrm{cm}$ above the sensors, and the responses of both the single-axis Hall array and the triaxial Hall sensor were recorded with and without the layer. For the single-axis array, the layer reduced the offset of the most affected channel from 244 to 180 ADC counts, a $26.2\%$ reduction. Although the number of affected channels increased from 4 to 13 out of 15, the interference was redistributed from a strong localized disturbance into a weaker, more spatially uniform bias, reducing large false-force responses in individual channels. For the triaxial sensor, the layer reduced the static-offset norm by $36.7\%$ and the dynamic RMS error by $33.2\%$.

\textbf{(4) Active compensation (FG-MBC).} To further mitigate magnetic interference, we next evaluated FG-MBC (Sec.~\ref{sec:fgmbc}) for active correction of the residual magnetic bias in the single-axis Hall array. Three randomly selected Hall channels were tested with the interfering magnet inclined at $45^{\circ}$ and positioned $1\,\mathrm{cm}$ from the sensor, as shown in Fig.~\ref{fig:forceTempCali}(b). Each channel underwent three loading cycles over $0$--$5\,\mathrm{N}$ under three conditions: clean, interfered without compensation, and interfered with FG-MBC enabled.

\begin{figure}[t]
    \centering
    \includegraphics[width=0.78\columnwidth]{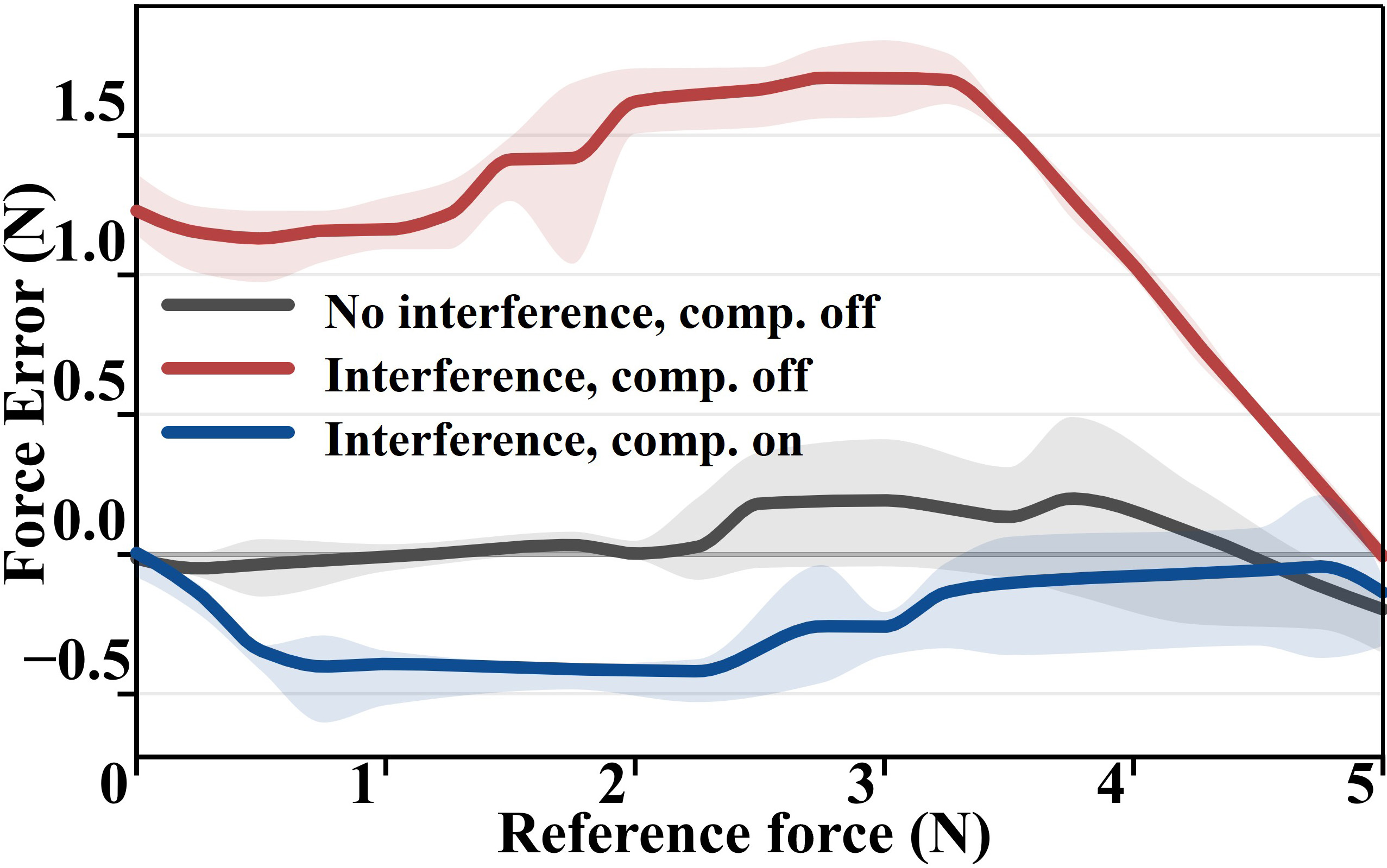}
\vspace{-4pt}
\caption{Average Hall-force error under three test conditions. Solid lines and shaded regions show the mean and min--max range over three trials, respectively. The red curve drops to zero at a reference force of $5\,\mathrm{N}$ because of sensor saturation.}
    \label{fig:compensation}
    \vspace{-18pt}
\end{figure}

As shown in Fig.~\ref{fig:compensation}, the mean RMSE increased from $0.23\pm0.15\,\mathrm{N}$ under clean conditions to $1.29\pm0.07\,\mathrm{N}$ under uncompensated interference. With FG-MBC, the RMSE decreased to $0.32\pm0.10\,\mathrm{N}$, approximately $75\%$ lower than the uncompensated value, while $R^{2}$ improved from 0.904 to 0.986. Together, these results show that passive shielding reduces localized magnetic interference and improves the robustness of the triaxial channel, while FG-MBC further suppresses the residual normal-force error to near the clean-condition level under magnetic disturbance.

\subsection{Post-Contact Stage: Thermal Material Recognition}
\label{sec:material}

\noindent Objects with similar geometry can be difficult to distinguish using ToF sensing alone. To address this limitation, thermal sensing provides complementary material information through
contact-induced heat transfer~\cite{bhattacharjee2015material,wade2017thermal}. We therefore evaluated whether thermal transients can distinguish objects with different surface materials.

\begin{figure}[t]
\centering
\includegraphics[width=0.99\columnwidth]{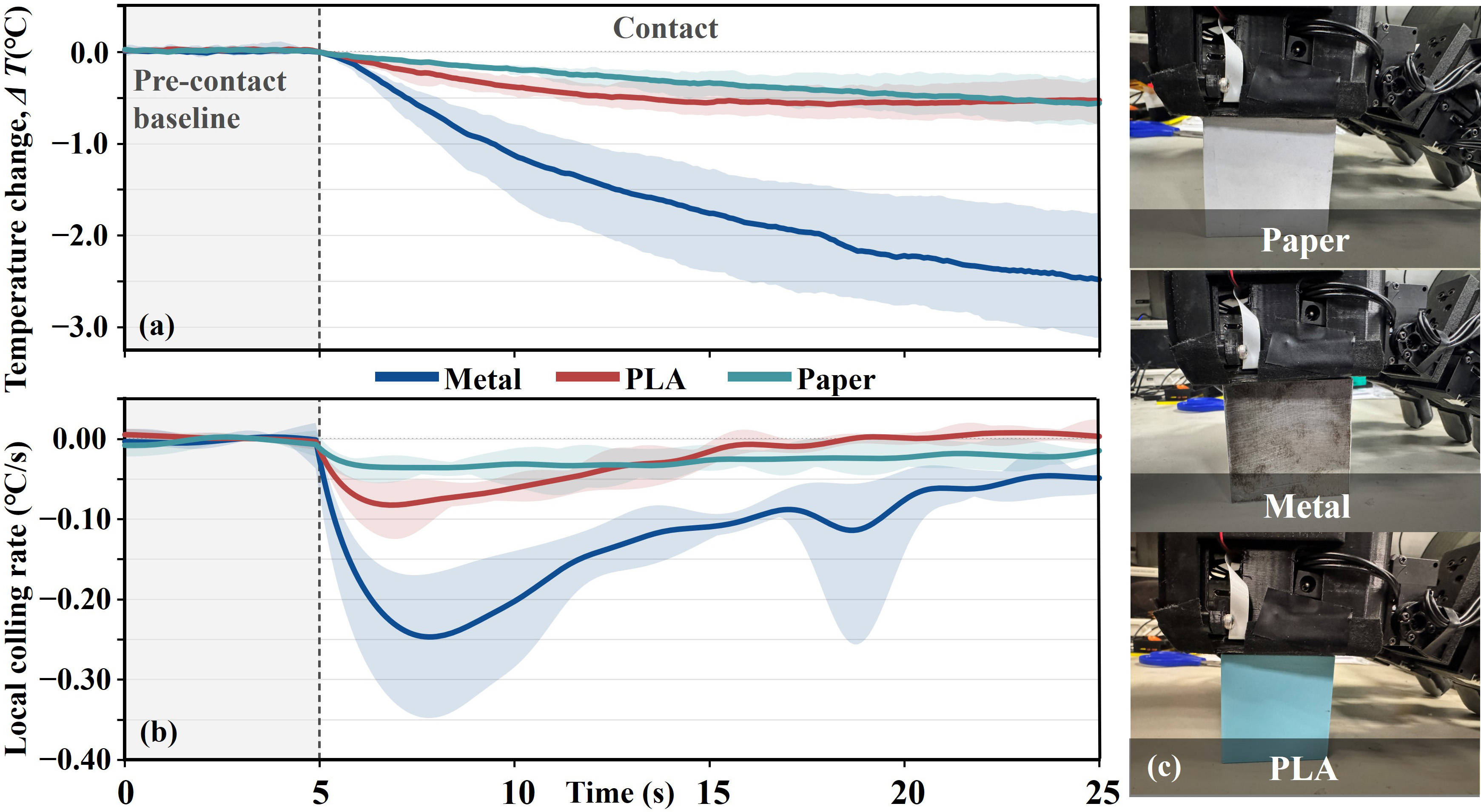}
\vspace{-15pt}
\caption{Thermal responses to metal, PLA, and paper. The dashed line marks contact onset.
\textbf{(a)} Temperature change relative to the temperature at contact onset.
\textbf{(b)} Rate of temperature change. Solid curves and shaded regions indicate trial means and ranges, respectively.
\textbf{(c)} Bio-Skin-Palm in contact with paper, metal, and PLA samples.}
\vspace{-18pt}
\label{fig:material}
\end{figure}

With Bio-Skin-Palm stabilized near its target temperature of $31^{\circ}\mathrm{C}$, the same palm region was pressed against each sample at a constant normal force of $3\,\mathrm{N}$ for $20\,\mathrm{s}$ (Fig.~\ref{fig:material}(c)). We tested objects with identical geometry but three different surface materials: metal, PLA, and paper. Each material was evaluated over five contact cycles at room temperature.

Fig.~\ref{fig:material}(a)(b) show clear material-dependent thermal responses. At $10\,\mathrm{s}$ after contact, the mean temperature changes were $-1.76\pm0.46$, $-0.55\pm0.13$, and $-0.33\pm0.11\,^{\circ}\mathrm{C}$ for metal, PLA, and paper, respectively. The corresponding peak cooling rates over the first $10\,\mathrm{s}$ were $-0.247\pm0.074$, $-0.084\pm0.025$, and $-0.053\pm0.013\,^{\circ}\mathrm{C\,s^{-1}}$. For both measures, all pairwise differences remained significant after Holm correction ($p_{\mathrm{adj}}\leq0.0317$, exact permutation test), indicating consistent material-dependent thermal responses. Although PLA and paper responses converge slower during contact, their early cooling rates remain well separated. We therefore use the peak cooling rate over the first $10\,\mathrm{s}$ as the material-discrimination cue, which also defines the contact duration used in Sec.~\ref{sec:closedloop}. These results demonstrate that thermal transients could provide material-dependent cues that complement the pre-contact geometric information.

\subsection{Staged Perception in the Loop: Autonomous Grasping}
\label{sec:closedloop}

\begin{figure*}[t]
    \centering
    \includegraphics[width=1\linewidth]{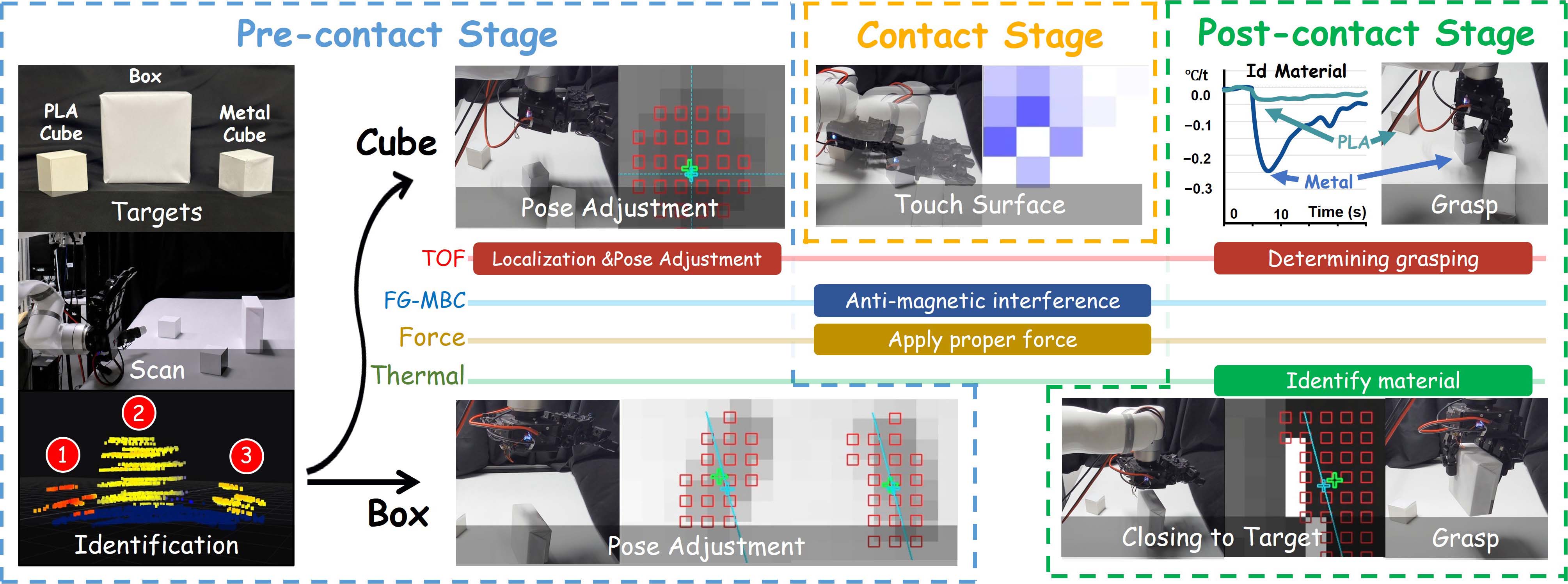}
    \caption{Closed-loop autonomous grasping. \textbf{Pre-Contact stage:} Reconstruct three tabletop objects from palm ToF measurements. ToF depth adjust the LEAP Hand pose with the target line. \textbf{Contacting stage:} For multiple candidates, the hand presses down after alignment until the sensed force reaches $3\,\mathrm{N}$ and holds for $10\,\mathrm{s}$ to record the temperature. \textbf{Post-Contact stage:} For a single box, the hand descends until the minimum ToF range falls below $3\,\mathrm{cm}$ before grasping. For multiple candidates, the temperature-change rate identifies the correct material, after which the arm moves above the target and grasps it.}
    \label{fig:pipeline}
    \vspace{-16pt}
\end{figure*}

\noindent We next evaluated how the complementary sensing modalities support autonomous, long-horizon manipulation without RGB input. Specifically, the system combines Bio-Skin-Palm with a locally deployed
vision--language model (VLM)~\cite{huang2023voxposer} to sequentially
localize the target, resolve material ambiguity, and execute the grasp (Fig.~\ref{fig:pipeline}).

\textbf{(1) Setup and protocol.}
We used an xArm with a LEAP Hand equipped with Bio-Skin-Palm to manipulate three tabletop objects: a $5\times15\times15\,\mathrm{cm}$ white box, a white metal cube, and a white PLA cube. Both cubes measured $60\,\mathrm{mm}$ per side and shared the same size, shape, and color, making their materials difficult to distinguish from geometry or color alone. Object positions, orientations, and left-to-right order were randomized across trials, with at least $10\,\mathrm{cm}$ separation between objects to reduce collision risk. We conducted 20 trials targeting the box and 10 trials targeting each cube material. For scene understanding, we locally deployed the Qwen3.6-35B-A3B VLM~\cite{qwen36_35b_a3b} using Q4\_K\_M quantization. %

\textbf{(2) Pre-contact: target localization and alignment.}
The xArm first followed a predefined scanning trajectory to acquire palm ToF measurements, which were reconstructed into a scene point cloud (Sec.~\ref{sec:pointcloud}). A front-view rendering of the point cloud was provided to the VLM to estimate the number of object clusters. This estimate guided adaptive DBSCAN~\cite{ester1996dbscan} to remove background and outlier points while retaining the estimated number of clusters.

The resulting clusters were labeled 1--3 from left to right, and their centroids were computed. The labeled rendering was then provided to the VLM together with the operator's natural-language request, which specified both geometry and material, such as ``rectangular box'' or ``small metal cube''. The VLM returned a single target index when geometry alone was sufficient, or two candidate indices when material discrimination was required.

For a single target, the arm moved to a position $30\,\mathrm{cm}$ above the target centroid for initial alignment. We then used an adaptive alignment strategy in which the live $8\times8$ ToF map and a reference image of the desired projected shape were provided to the VLM. The system adjusted the $xyz$ translation and wrist rotation in small increments, sending the updated ToF map to the VLM after each movement until the observed shape aligned with the reference image. For two candidates, the arm visited each candidate in turn and used ToF feedback to center it beneath the palm before contact.

\begin{table}[t]
\centering
\caption{Performance of closed-loop autonomous grasping.}
\vspace{-4pt}
\label{tab:closedloop}
\small
\setlength{\tabcolsep}{4pt}
\renewcommand{\arraystretch}{1.1}
\begin{tabular}{llcc}
\toprule
Track & Evaluation item & Succ./Total & Rate \\
\midrule
\multirow{3}{*}{Box}
 & Cluster count correctly                   & 20/20 & 100\% \\
 & Box correctly selected                    & 19/20 & 95.0\% \\
 & \textbf{End-to-end}                       & \textbf{18/20} & \textbf{90.0\%} \\
\midrule
\multirow{4}{*}{Cube}
 & Cluster count correctly                   & 20/20 & 100\% \\
 & Both cubes correctly selected             & 17/20 & 85.0\% \\
 & Material correctly Identified             & 16/17 & 94.1\% \\
 & \textbf{End-to-end}                       & \textbf{16/20} & \textbf{80.0\%} \\
\bottomrule
\end{tabular}
\vspace{-16pt}
\end{table}

\textbf{(3) Contact: force-controlled thermal measurement.}
The system then used contact-induced thermal responses to discriminate between materials. Specifically, the palm descended onto each candidate until any Hall channel reached $3\,\mathrm{N}$, establishing a consistent contact force across trials. Contact was maintained for $10\,\mathrm{s}$ while the palm surface temperature was recorded.

\textbf{(4) Post-contact: material discrimination and grasping.}
For the box, geometry alone was sufficient for target selection. In contrast, the two cubes had identical geometry and therefore required material discrimination. The system computed the peak cooling rate during the first $10\,\mathrm{s}$ of contact. The candidate with the larger absolute cooling rate was classified as metal, while the other was classified as PLA. The arm then returned to the selected target and performed ToF-based alignment.

After alignment, the arm descended until the minimum ToF range fell below $3\,\mathrm{cm}$. It then stopped, closed the hand, and attempted to lift the object by $30\,\mathrm{cm}$. A lift was classified as stable if no ToF return decreased by more than $5\%$ during the lift. Because the metal cube exceeded the payload limit of the LEAP Hand, trials targeting the cube were considered successful when the hand closed on the correct cube.

\textbf{(5) Results.} As shown in Table~\ref{tab:closedloop}, the system succeeded in 18 of 20 box trials (90.0\%) and 16 of 20 cube trials (80.0\%). The VLM correctly counted the point-cloud clusters in all 40 trials. The two box failures resulted from incorrect target selection and an unsuccessful grasp, respectively. In three cube trials, the VLM confused boxes with cubes, causing the system to skip material recognition. Of the remaining 17 trials, thermal material recognition succeeded in 16 (94.1\%). Overall, the high success rates demonstrate the effectiveness of the integrated system.

\section{Conclusion}
\label{sec:conclusion}

\noindent We presented Bio-Skin-Palm, an open-source, $5.1\,\mathrm{mm}$-thick multimodal tactile palm with a cost of approximately 30\,USD. Its layered design integrates distributed normal-force sensing, global shear sensing, proximity sensing, and active thermal regulation, enabling tactile perception beyond fingertip contact across pre-contact geometry, contact forces, and post-contact material cues. A hybrid strategy combining passive permalloy shielding with active FG-MBC compensation further improves robustness to external magnetic interference. Through evaluations on an autonomous multi-target task, we demonstrated both the capabilities of individual modalities and their complementary roles in manipulation without RGB-camera input.

Two limitations remain. First, FG-MBC corrects only normal-force interference, while shear relies on passive shielding. It also cannot track interference that varies during contact because the bias estimate is fixed once contact is detected. In such cases, the magnetically insensitive FSR array can provide a normal-force fallback to avoid large interference-induced errors, albeit at lower accuracy. Second, thermal material recognition is sensitive to the effective contact area. Future work will investigate automatic switching to the FSR fallback when Hall--FSR measurements disagree, learned material classifiers trained on larger and more diverse multimodal datasets, and applications with vision-language-action (VLA) and world action models (WAMs).

\appendix
\section{Calibration and Implementation Details}

\subsection{Force Calibration}
\label{app:forcecal}
\noindent \textbf{Normal force.}
Hall and FSR responses are calibrated at thermal steady state using separate loading and unloading curves, whose difference enables the calculation of hysteresis.

\noindent \textbf{Shear force.}
The baseline-subtracted triaxial Hall field $\Delta\mathbf B$ is mapped to the planar shear force $\mathbf F_{xy}$ by polynomial ridge regression:
\begin{equation}
\mathbf W^{*}
=
\arg\min_{\mathbf W}
\sum_n
\left\|
\mathbf W\boldsymbol{\phi}(\Delta\mathbf B_n)
-\mathbf F_{xy,n}
\right\|_2^2
+\lambda\|\mathbf W\|_F^2 .
\end{equation}
Here, $\boldsymbol{\phi}(\cdot)$ is a degree-$d$ polynomial feature map, and $\lambda$ is the ridge weight. Leave-one-cycle-out cross-validation gives $d=4$, $M=34$, and $\lambda=10$. The estimated shear force is
$\hat{\mathbf F}_{xy}=\mathbf W^{*}\boldsymbol{\phi}(\Delta\mathbf B)$.
Calibration and evaluation in the four cardinal directions are with 0.5N preload.

\subsection{Thermal Calibration and Feature Extraction}
\label{app:thermalcal}

\noindent The NTC sensors are calibrated against thermal-camera measurements. During testing, the palm temperature is maintained at approximately $31^{\circ}\mathrm{C}$ in a $24^{\circ}\mathrm{C}$ environment. Material recognition uses the peak cooling rate $r_{\min}$ within $10\,\mathrm{s}$ after contact onset $t_0$, defined as the minimum temperature derivative estimated by causal local linear fitting.

\subsection{FG-MBC Parameters}
\label{app:fgmbc}

\noindent FG-MBC operates with a frame interval of $\Delta t=2\,\mathrm{ms}$. The adaptation and decay time constants are $\tau_a=0.006\,\mathrm{s}$ and $\tau_r=0.070\,\mathrm{s}$, respectively, yielding $\alpha_{a,r}=1-\exp(-\Delta t/\tau_{a,r})$ in (\ref{eq:fgmbc}). The detection signal is low-pass filtered with a time constant of $0.025\,\mathrm{s}$. Since FSR zones fewer than Hall channels, one FSR zone is enlarged to cover four adjacent Hall channels.

\bibliographystyle{IEEEtran}
\bibliography{references}

\end{document}